\documentclass[10pt,twocolumn,letterpaper]{article}

\usepackage[pagenumbers]{cvpr} 
\usepackage{booktabs,tabularx}
\definecolor{cvprblue}{rgb}{0.21,0.49,0.74}
\usepackage[pagebackref,breaklinks,colorlinks,allcolors=cvprblue]{hyperref}

\def\paperID{4112} 
\def\confName{CVPR}
\def\confYear{2025}

\title{ActNAS : Generating Efficient YOLO Models using Activation NAS}

\author{
  {\bf Sudhakar Sah, Ravish Kumar, Darshan C. Ganji, Ehsan Saboori} \\ \\
  Deeplite, Inc.\\
  Toronto, Canada \\
  \texttt{sudhakar@deeplite.ai}
}

\begin{document}
\maketitle
\begin{abstract}
Activation functions introduce non-linearity into neural networks, allowing them to learn complex patterns. Different activation functions impact both speed and accuracy; for instance, ReLU is fast but often less precise, while SiLU offers higher accuracy at the expense of speed. Traditionally, a single activation function is used across the entire model. In this work, we conducted a comprehensive study on the effects of using mixed activation functions in YOLO-based models, examining their impact on latency, memory usage, and accuracy across CPU, NPU, and GPU edge devices. We propose a novel approach, Activation NAS (ActNAS), which utilizes Hardware-Aware Neural Architecture Search (HA-NAS) to design YOLO models with mixed activation functions optimized for specific hardware. Models generated through ActNAS achieve similar mean Average Precision (mAP) compared to baseline models, while running up to 1.67 times faster and/or using 64.15\% less memory on target devices. Additionally, we demonstrate that hardware-aware models learn to leverage architectural and compiler-level optimizations, resulting in highly efficient performance tailored to each hardware platform.
\end{abstract}    
\section{Introduction}
\label{sec:intro}
\begin{figure}
    \centering
    \includegraphics[width=1.0\linewidth]{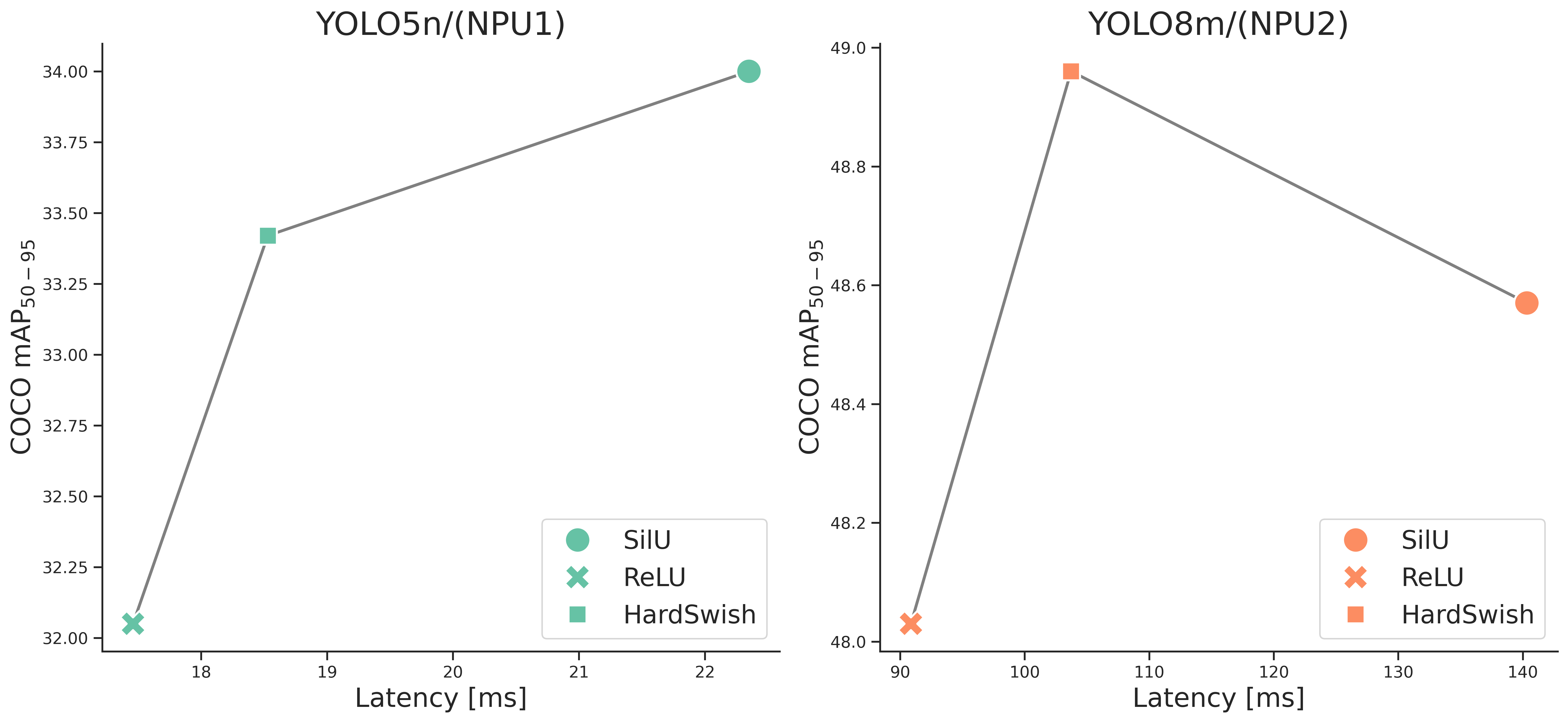}
    \caption{Latency-mAP plot of YOLO5n and YOLO8m for two different NPUs. SiLU models are slowest and ReLU models are fastest on both devices}
    \label{fig:act_ref}
\end{figure}

Improving the accuracy of computer vision models has become a highly competitive area, with researchers constantly refining or developing new architectures to achieve state-of-the-art performance. For example, the recent YOLO10 \cite{YOLOv10} architecture has set a new benchmark for object detection within the YOLO family \cite{yolo_survey}, outperforming previous models in the popular Microsoft COCO dataset \cite{lin2014microsoft}. The current trend toward higher accuracy often involves designing more complex models by increasing the number of learnable parameters or adding performance-enhancing blocks. However, these gains in accuracy typically come at the cost of higher latency. Some research works redesign activation functions to learn complex patterns and improve accuracy-latency tradeoffs. Since these activation functions are applied throughout the model architecture, they significantly influence both model latency and accuracy. For example, the Sigmoid Linear Unit (SiLU) \cite{silu} offers better accuracy at the expense of higher latency (slower) than the less accurate but less latent (faster) Rectified Linear Unit (ReLU) \cite{relu} and HardSwish \cite{hardswish} activation function offering the best latency-accuracy tradeoff. Figure \ref{fig:act_ref} demonstrates the accuracy-latency tradeoff of these variations for YOLO5n and YOLO8m models.
A common practice in deep learning model design is using a single activation function across all layers in the model architecture. Through our exploration, we observed that employing different activation functions for various layers in the model either improves the mean Average Precision (mAP) by less than 1\% or stays within 1\% drop but reduces latency by 30-70\%, compared to the baseline model using SiLU activation across all layers. In this study, we controlled all other factors that could affect latency and mean average precision (mAP) to isolate the impact of the mixed activation function design. We also propose a Hardware Aware Neural Architecture Search (HA-NAS) approach to automatically select the optimal activation function for each layer based on given latency and accuracy constraints.\\

We summarize our contributions as follows.
\begin{itemize}
    \item We introduce {\em ActNAS}, a Hardware Aware NAS based method to search for the best configuration of the model using a mixed activation function for each layer.
    \item We summarize the impact of different activation functions on memory and latency on different edge devices. 
    \item We demonstrate that our ActNAS generated models have
    negligible drop in mAP, but significant improvement 
    over latency compared to the baseline models trained on COCO dataset and tested on difference edge devices.
    \item We also demonstrate the hardware aware model for one device may not necessarily perform well on other type of device.
\end{itemize}
\section{Related Works}
\label{related_work}
In building computer vision models, it is a common practice to use the same activation function across all layers of the architecture. For example, YOLO models \cite{yolo_survey}, whether in the n, m, s, l, or tiny variants, consistently use the SiLU \cite{silu} activation function due to its superior accuracy compared to alternatives like ReLU \cite{search_activation}. However, this higher accuracy often comes at the cost of increased latency, which opens up the possibility of exploring other activation functions such as ReLU \cite{relu}, LeakyReLU\cite {leakyrelu}, or Hardswish \cite {hardswish}. While YOLO models do allow the option to switch from SiLU to other activations across the entire architecture, there has been no prior research (based on our knowledge) on using mixed activation functions within the same YOLO model. The closest existing work involves designing hybrid activation functions, simplifying blocks in the architecture by applying a different activation function than the main one or mixing activations in image classification models.

\textbf{Hybrid Activation}: In \cite{Hybrid_Activation}, the authors combined softmax \cite{softmax} and sparsemax\cite {sparsemax} in the final activation layer of a Convolutional Neural Network (CNN) for gait analysis using silhouettes. Similarly, in \cite{Adaptive_hybrid_activation}, the authors proposed an activation function that can replace ReLU, SiLU, and Hardswish in deep learning models, providing a hybrid approach.

\textbf{Block Simplification}:
YOLOv6-3.0 \cite{yolov6} introduced the idea of simplifying the neck of YOLO models \cite{yolo_survey} by replacing the SiLU activation in the SPPF block \cite{yolov6_ultralyticsv5} with ReLU, creating the SimSPPF block. This was further modified into the SimCSPSPPF block to enhance performance. These works demonstrate that different activation functions have unique strengths and weaknesses, motivating us to explore mixed activations within a single YOLO architecture.

\textbf{Designing Model Architecture using NAS}: In \cite{NAS-original}, Zoph demonstrated how to design Recurrent Neural Networks (RNNs) with constituent nodes using different activation functions such as ReLU, identity, tanh, or sigmoid. Building on this, \cite{NAS-Pyramidal-Topology} introduced a method to search across activation functions like ReLU and Gaussian-smoothed ReLU. Zhenyu \cite{NAS-skip-connection} further proposed Eigen-NAS, a train-free algorithm that searches for optimal skip connections and activations for each layer in image classification models, showing improvements on datasets like CIFAR-100 and ImageNet-16 but observed a drop in accuracy with CIFAR-10. Our work extends this approach of mixing activations to YOLO architectures by proposing ActNAS, a solution that optimizes activation functions within YOLO models based on layer-specific accuracy and latency data.

\textbf{Zero-Cost (ZC) Estimators}: Zero-cost estimators, which provide accuracy predictions without full training, have been shown to significantly reduce training time. In \cite{yolobench}, Ivan et al. analyzed various ZC estimators, including those from \cite{mellor2021neural, abdelfattah2021zero, li2023zico}, in the context of YOLO models, concluding that the NWOT metric \cite{mellor2021neural} performs well for networks using different activations. However, ZC estimators have not yet been applied to assess the micro-level impact of architectural changes, such as altering individual activation functions within a model.

\begin{figure}
    \centering
    \includegraphics[width=1.0\linewidth]{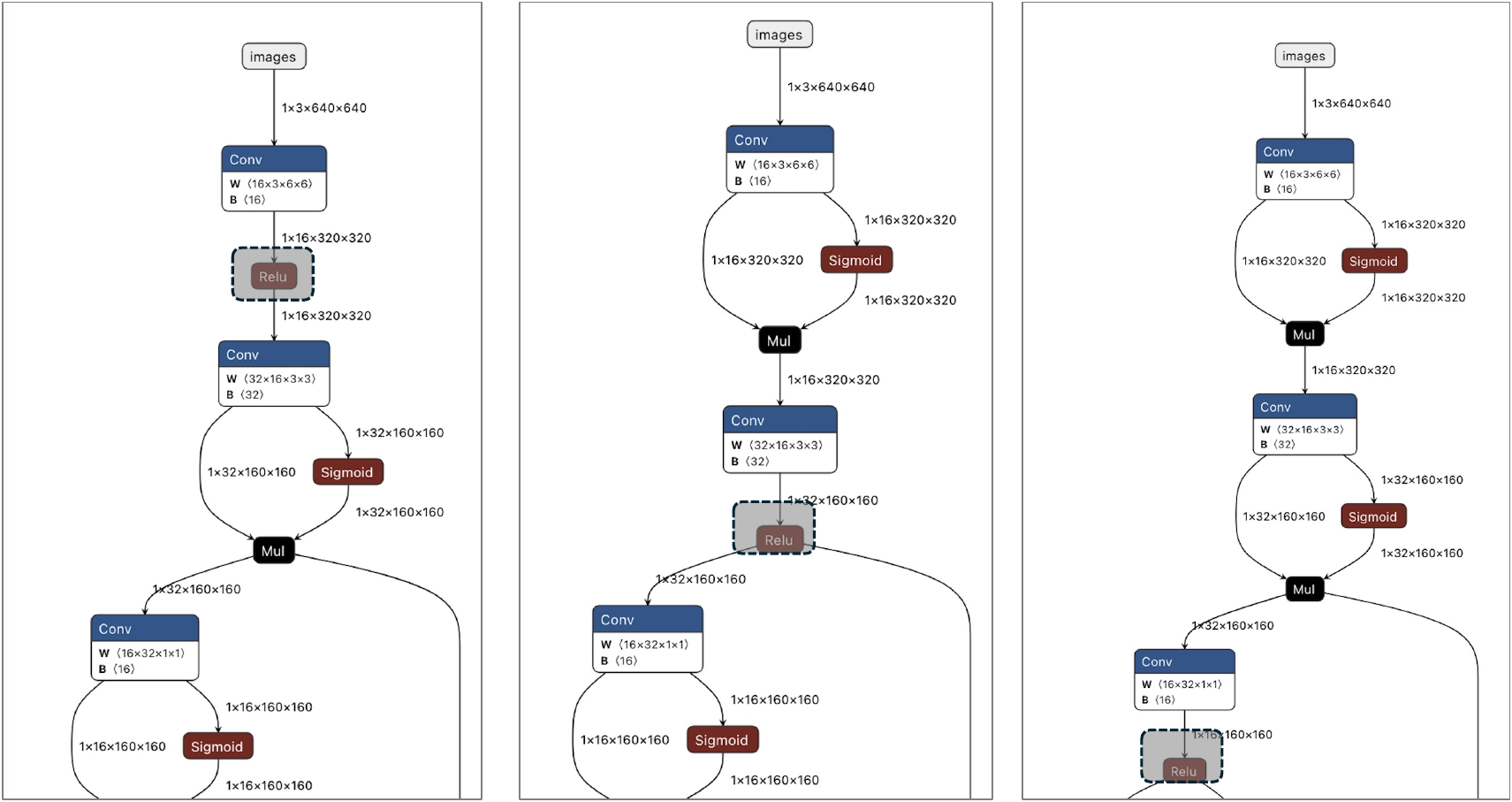}
    \caption{Layer wise activation replacement a) Replace first SiLU activation to ReLU b) Replace second activation c) Replace third activation}
    \label{fig:layer_wise_replacement}
\end{figure}
\section{Methodology}
\label{methodology}

Unlike the straightforward approach of using the same activation function across all layers in a model, mixed activation models incorporate multiple activation functions within the architecture. As illustrated in Figure \ref{fig:act_ref}, the YOLO5n and YOLO8m models with SiLU activation achieve higher mAP but is least efficient on different devices, while the model with ReLU activation is the fastest with lower mAP. The goal of mixed activation models is to identify the optimal combination of activation functions for each layer, creating a model that strikes the best balance between latency and accuracy.

\begin{figure*}
    \centering
    \includegraphics[width=1.0\linewidth]
    {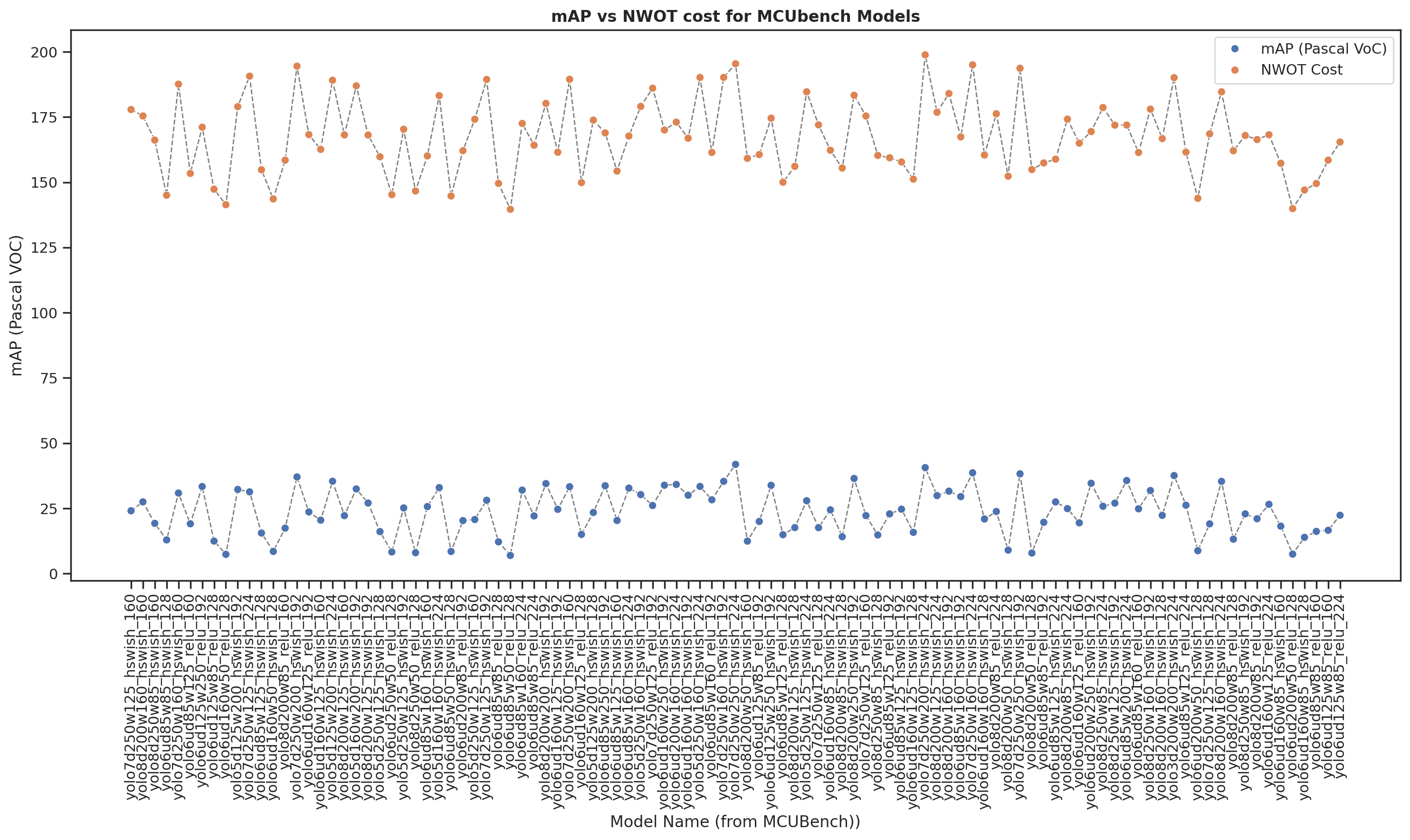}
    \caption{mAP vs NWOT cost for 131 MCUbench\cite{sah2024mcubench} Models. Each point represents one model and NWOT and mAP score of each model is plotted in orange and blue colors respectively}
    \label{fig:model_zc_metric}
\end{figure*}

\begin{figure*}
    \centering
    \includegraphics[width=1.0\linewidth]{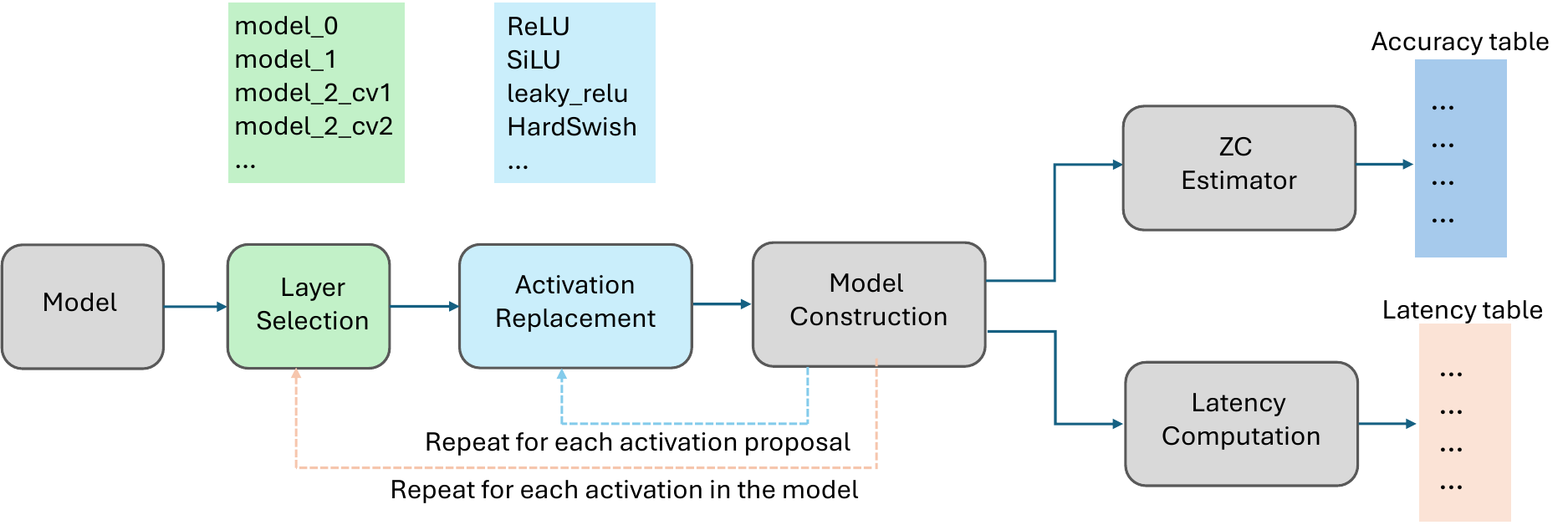}
    \caption{Model Benchmarking using training free estimators and on device inference. Each candidate model is created by replacing just one activation function. }
    \label{fig:model_benchmarking}
\end{figure*}

For our experiments, we used YOLO5n combined with YOLO8 head (which offers better accuracy) and YOLO8m model, with the SiLU-based model serving as a reference. As shown in Figure \ref{fig:model_benchmarking}, we generate a search space by systematically replacing each activation function in the reference model with a set of candidate activations: \textit{[ReLU, SiLU, Hardswish, ReLU6, and LeakyReLU]}. We replace one activation at a time, as depicted in Figure \ref{fig:layer_wise_replacement}, resulting in 345 candidate models for YOLO5n (since the reference model has 69 activations and there are 5 candidate activations for each layer).

To evaluate the impact of each activation replacement, we create a new model for each candidate and measure its accuracy, latency, and memory usage, comparing it to the reference model. This process is iterative and is repeated for all 345 candidate models. The results are recorded in performance tables: the latency table, accuracy table, and memory table. Each entry in the accuracy table includes the layer name, activation name, reference accuracy, and delta accuracy (the difference between the reference model's accuracy and the candidate model). The latency and memory tables follow the same process and record latency and memory values instead of accuracy.

\subsection{Accuracy Estimator}
As the number of activations and layer combinations increase, the search space grows exponentially, making it computationally expensive and time-consuming to train each model and evaluate the impact of replacing activations. To address this, we use the NWOT Zero-Cost (ZC) metric \cite{mellor2021neural} to approximate the impact of changing activation functions in different layers on the accuracy of the model.

To validate the effectiveness of  NWOT score as a proxy to the mAP of trained YOLO models, we estimated the correlation between the mAP of 131 variations of YOLO models in MCUBench \cite{sah2024mcubench} and their corresponding NWOT scores. Figure \ref{fig:model_zc_metric} shows that the  NWOT score has a strong correlation with the mAP of fully trained models. The high correlation indicates that the NWOT score can reliably predict the impact of per-layer activation changes on accuracy without the need of training each candidate model. For each candidate activation replacement, the NWOT score is calculated, and the accuracy table is updated accordingly as described in the previous section. This approach significantly reduces the computational cost and time required to explore the large search space of mixed activation models. The suitability of zero cost proxy based method also opens an avenues for more complex NAS methods requiring larger search space. 

\subsection{Latency Computation}

The impact of different activation functions on latency and memory depends on the hardware architecture, compiler level optimization and the runtime.
For this reason, a variety of hardware, including two CPUs (ARM Cortex-A57 on Jetson Nano and ARM Cortex-A53
on Raspberry Pi 3), an embedded GPU (Jetson Nano), and three different reference neural processing units (NPU1, NPU2, NPU3) are used for our experiments. We are not disclosing the vendor name and actual name of these NPUs as some of these are yet to available in the market. To measure the performance of the models on these devices, we used ONNX Runtime \cite{onnx} for Jetson Nano, TensorFlow Lite (TFLite) \cite{tflite} for ARM Cortex processors, and custom compilers/runtimes as needed by each NPU. Each model is converted to ONNX, TFLite, or a custom format depending on the compiler. Latency values are recorded as average over 50 runs to ensure correctness. The input size of $224\times224$ or $640\times640$ is used depending upon the memory limitations. This variation in hardware, runtime and model family ensures that the results aren't biased towards a specific platform, runtime or model family, providing a balanced perspective of how different activations perform across different systems.

\subsection{Mixed Activation Model Search}
Latency, memory and accuracy table obtained from the benchmarking step is used to construct mixed activation models. 

\subsubsection{Local Zero Cost Maxima Approach}
We started with the simple approach to create a model using just two activation functions, such as SiLU and ReLU. Starting with a reference model that uses only SiLU, we apply the Local Zero Cost Maxima (LZCM) method to iteratively replace activation of a layer with ReLU provided the NWOT score is higher compared to the model that uses SiLU for the same layer. The modified model is then trained from scratch on the COCO dataset. Using this simple approach, we developed two models: LZCM1, by using SiLU and ReLU, and LZCM2, by using Hardswish and ReLU activations.

\begin{figure*}
    \centering
    \includegraphics[width=1.0\linewidth]{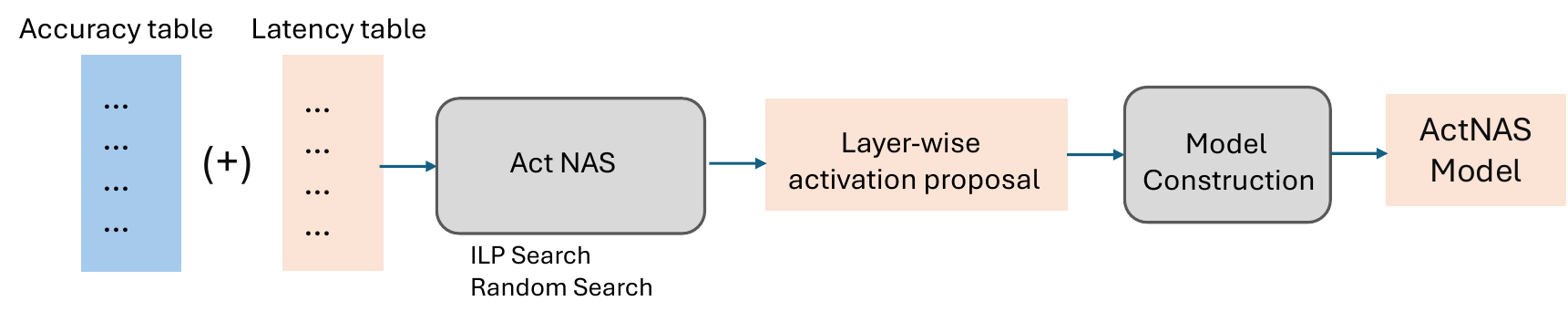}
    \caption{Activation NAS Process}
    \label{fig:activaton_nas_process}
\end{figure*}

\subsubsection{Naive Approach}
Benchmarking models generated using the LZCM approach show that activations in layers closer to input have a significant impact on both latency and memory due to the larger feature sizes in those layers. Replacement of activations in initial layers with ReLU significantly reduces latency and memory usage. In our simple approach, we replace the first three activations near the input with ReLU, while keeping the remaining layers as SiLU. This leads to an efficient balance between performance and resource consumption. 

\subsubsection{Activation NAS Approach}
The LZCM method follows a straightforward decision-making process, replacing activations in individual layers according to their zero-cost score. However, this approach only optimizes each layer locally and does not account for the combined impact of replacing activations in multiple layers at once. In contrast, the Activation NAS (ActNAS) approach searches for the optimal activation function for each layer while considering a global budget for latency, memory, or accuracy. The search space for ActNAS is defined by the number of layers multiplied by the number of activation function options. To navigate this space, we used the latency, memory, and accuracy tables, along with predefined constraints, to guide the search. We experimented with both random search and Integer Linear Programming (ILP) \cite{trauth1969integerlinear} to find the best activation configurations.

\paragraph{Random Search}
As shown in Figure \ref{fig:activaton_nas_process}, generating model proposals using Activation NAS requires precomputed latency and accuracy tables. These tables capture the changes in these metrics when a single layer's activation in the original YOLO model (with SiLU activation) is replaced by an alternative activation. The tables are then converted into latency and accuracy cost matrices, where the rows represent the layers being modified, and the columns represent candidate activations (SiLU, ReLU, Hardswish, LeakyReLU and ReLU6). In addition to these matrices, we also define latency or accuracy constraints, which serve as the overall budget that the algorithm can work within when constructing new models with mixed activations.
The random search process begins by randomly selecting activations from the search space that fit within the given constraints. Then it computes the total cost (in terms of latency or accuracy) for the model and marks it as the best possible model at that point. The algorithm repeats the random search, comparing each newly constructed model with the previous best. If a new model has a better overall cost, it updates the best model definition and continues iterating through the random selection process.

\begin{table*}[ht!]
\caption{Performance of ActNAS models compared to baseline YOLO5n models on GPU and CPU}
\centering
\setlength\tabcolsep{10pt}
\label{table:table_cpu_gpu}
\begin{tabularx}{\linewidth}{p{3.3cm}|m{1.2cm}|ccc|ccc}
\toprule
\textbf{Model} & \textbf{mAP} & \multicolumn{3}{c}{\textbf{Jetson Nano}} & \multicolumn{3}{c}{\textbf{Cortex A-53}} \\
\midrule
 & & Latency & \multicolumn{2}{c}{Improvement (\%)} & Latency & \multicolumn{2}{c}{Improvement (\%)} \\
\midrule 
& & (ms) & Hswish & SiLU & (ms) & Hswish & SiLU \\
\midrule

SiLU & \textbf{33.60} & \textbf{27.20} & {-} & {-} & 937.89          & {-} & {-} \\ 
ReLU          & \textbf{31.88} & \textbf{20.18} & {-} & {-} & \textbf{793.27} & {-} & {-} \\
Hardswish          & \textbf{33.13} & \textbf{25.63} & {-} & {-} & \textbf{823.91} & {-} & {-} \\
LZCM1(SiLU/ReLU)           & 33.13          & 26.61          & -21.54\%              & 2.17\%                & 916.97          & -0.58\%               & 2.23\%                \\
LZCM2(SilU/Hswish)           & 33.29          & 31.15          & -21.54\%              & -14.52\%              & 828.72          & -0.58\%               & 11.64\%               \\
Naive(ReLU/SiLU)           & 33.52          & 24.55          & 4.21\%                & 9.74\%                & 906.78          & -10.06\%              & 3.32\%                \\
ActNAS5n.1          & \textbf{32.80} & \textbf{24.78} & \textbf{3.32\%}       & \textbf{8.90\%}       & \textbf{809.96} & \textbf{1.69\%}       & \textbf{13.64\%}      \\
ActNAS5n.2          & 32.20          & 23.50          & \textbf{8.31\%}       & \textbf{13.60\%}      & 966.74          & {-17.34\%}     & {-3.08\%}      \\
\bottomrule
\end{tabularx}

\end{table*}

\begin{table*}
\caption{Performance of ActNAS models compared to baseline models on NPU}
\label{table:table_npu}
\centering
\setlength\tabcolsep{10pt}
\begin{tabularx}{\linewidth}{p{3.3cm}|m{1.2cm}|ccc|ccc}
\toprule
\textbf{Model} & \textbf{mAP} & \multicolumn{3}{c}{\textbf{NPU1}} & \multicolumn{3}{c}{\textbf{NPU1}} \\
\midrule
 & & Latency & \multicolumn{2}{c}{Improvement (\%)} & RAM & \multicolumn{2}{c}{Improvement (\%)} \\
\midrule
& & (ms) & Hswish & SiLU & (KB) & Hswish & SiLU \\
\midrule

SiLU & \textbf{0.3366} & \textbf{22.35} & {-} & {-} & 1230.00 & {-} & {-} \\
ReLU & \textbf{0.3188} & \textbf{17.46} & {-} & {-} & \textbf{588.00} & {-} & {-} \\
Hardswish & \textbf{0.3313} & \textbf{18.53} & {-} & {-} & \textbf{392.00} & {-} & {-}   \\
LZCM1(SiLU/ReLU)  & 0.3313          & 21.87          & -5.50\%               & 2.15\%                & 1200.00         & -206.12\%             & 2.44\%                \\
LZCM2(SiLU/Hswish)  & 0.3329          & 19.55          & -5.50\%               & 12.53\%               & 624.75          & -59.38\%              & 49.21\%               \\
Naive(SiLU/ReLU)  & 0.3352          & 21.43          & -15.65\%              & 4.12\%                & 661.50          & -68.75\%              & 46.22\%               \\
ActNAS5n.1 & \textbf{0.3280} & \textbf{17.37} & \textbf{6.26\%}       & \textbf{22.28\%}      & 514.50          & {-31.25\%}     & \textbf{58.17\%}      \\
ActNAS5n.2 & \textbf{0.3250} & \textbf{17.50} & \textbf{5.56\%}       & \textbf{21.70\%}      & \textbf{441.00} & {-12.50\%}     & \textbf{64.15\%}\\
\bottomrule
\end{tabularx}

\end{table*}

\paragraph{ILP Search}
To generate model proposals with mixed activations, we first created latency and accuracy matrices, where each element represents the difference in metrics from the original SiLU-based YOLO model when a single-layer activation is replaced. We then formulated this problem as an Integer Linear Programming (ILP) \cite{trauth1969integerlinear} optimization problem. The objective was to minimize the cost of latency and accuracy while adhering to the constraints on the overall latency or accuracy of the newly constructed models. In the ILP formulation, one variable vector represents the indices of the layers where activations are applied, and the other represents the candidate activations for those layers. We used the open-source PuLP library \cite{PuLP}, a Python toolkit for linear programming, to solve this ILP problem. The result is a ranking list of the top \textit{k} model proposals. To ensure diversity among proposals, the solution avoids excessive overlap, keeping the model architectures meaningfully different.

The ActNAS models (ActNAS5n.1 is NPU1 aware and ActNAS5n.2 is Jetson Nano aware) in Table \ref{table:table_cpu_gpu} and \ref{table:table_npu} are hardware aware mixed activation models (YOLO5n) which means that we used the latency of each candidate models on NPU1 to create the optimized models. Similarly, ActNAS8m.1 and ActNAS8m.2 models in Table \ref{table:table_npus_yolov8} are NPU2 aware mixed activation models (YOLO8m). The ILP search can be adapted for different hardware targets by imposing a latency constraint on the constructed model. 

\subsection{Model Fine-tuning}
We used Ultralytics \cite{Ultralytics} to train and fine-tune all our models. For consistency across experiments, we kept all hyperparameters exactly the same, training each model from scratch for 300 epochs using $640\times640$ COCO images and evaluating them on the COCO validation set which contains $5000$ images. While mixed activation models could be fine-tuned for a few epochs starting from pre-trained weights, we chose to train all models from scratch to ensure a fair comparison.

\begin{table}[ht!]
\caption{Hardware aware performance of YOLO8m ActNAS models on different NPUs. All latency numbers are for $640\times640$ images. NPU1 is a low end NPU.}
\centering
\setlength\tabcolsep{9pt}
\label{table:table_npus_yolov8}
\begin{tabularx}{\linewidth}{p{1.7cm}|c|m{0.80cm}|m{0.8cm}|m{0.8cm}}
\toprule
\textbf{Model} & \textbf{mAP} & \multicolumn{3}{c}{\textbf{Latency(ms)}} \\
\midrule
 & & {NPU1} & {NPU2} & {NPU3} \\
  & & {x100} & {} & {} \\
\midrule

SiLU & 48.57 & 241.50 & \textbf{190.87} & 340.00  \\
HardSwish &  \textbf{48.96} & \textbf{166.80} & \textbf{140.33} & 308.80  \\
ReLU & 48.03 & \text93.17 & 103.73 & 283.40  \\
ActNAS8m.1 & 48.50 & 188.40 &  136.85 & 319.60  \\
ActNAS8m.2 & \textbf{48.60} & \textbf{143.80} & \textbf{114.33} & 305.48  \\

\bottomrule
\end{tabularx}
\end{table}

\section{Results}
\label{results}

We conducted our experiments on YOLO5n and YOLO8m \cite{Ultralytics} models and the selection is to cover two family of YOLO models with different levels of complexity. Initial benchmarks were performed on YOLO5n model for Pascal VOC dataset \cite{pascal-voc-2012} to avoid training multiple models on the larger COCO dataset \cite{lin2014microsoft}. Still, all the final benchmarking is performed on COCO dataset. Tables \ref{table:table_cpu_gpu} and \ref{table:table_npu}  summarize mAP, RAM (NPU only), and latency values for both the reference models and our mixed activation models for YOLO5n model. The baseline models consist of COCO-trained YOLO5 using SiLU, ReLU, and Hardswish activations, respectively. The LZCM1 model aims to minimize overall latency by mixing SiLU and ReLU activations using a simple search approach, while the LZCM2 model prioritizes maximizing accuracy by combining the more accurate activations, SiLU and Hardswish. Both models achieve mAP values slightly lower, but very close to SiLU and Hardswish reference models. In terms of latency, the LZCM1 model performs slightly better than the SiLU model across all devices (e.g., 21.87 ms vs. 22.35 ms on the NPU). The SiLU/Hardswish mixed activation model demonstrates a better accuracy-latency trade-off, being 12.5\% faster than the SiLU model but 5.3\% slower than the Hardswish model on the NPU. A similar trend is observed on A53 and A57 CPUs, as well as the Jetson Nano GPU. Lastly, the naive approach of replacing the first three activations from SiLU to ReLU resulted in slightly slower performance compared to other baseline models, but the drop in mAP was less than 1\% compared to the SiLU model.

\begin{figure*}
    \centering
    \includegraphics[width=0.8\linewidth]{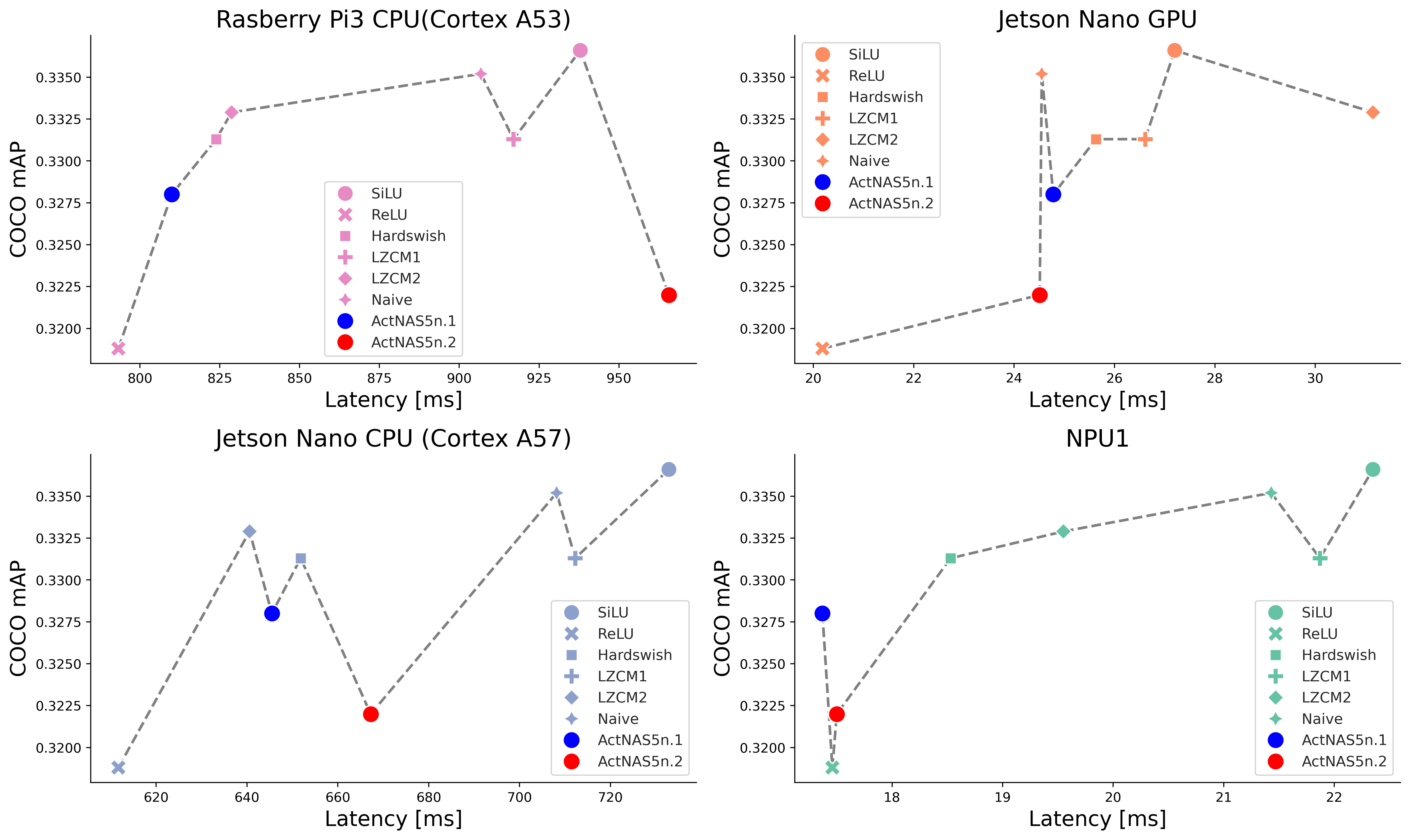}
    \caption{Performance of YOLO5n reference models and ActNAS models on different edge devices.}
    \label{fig:results_plot}
\end{figure*}

\begin{figure*}
    \centering
    \includegraphics[width=1.0\linewidth]{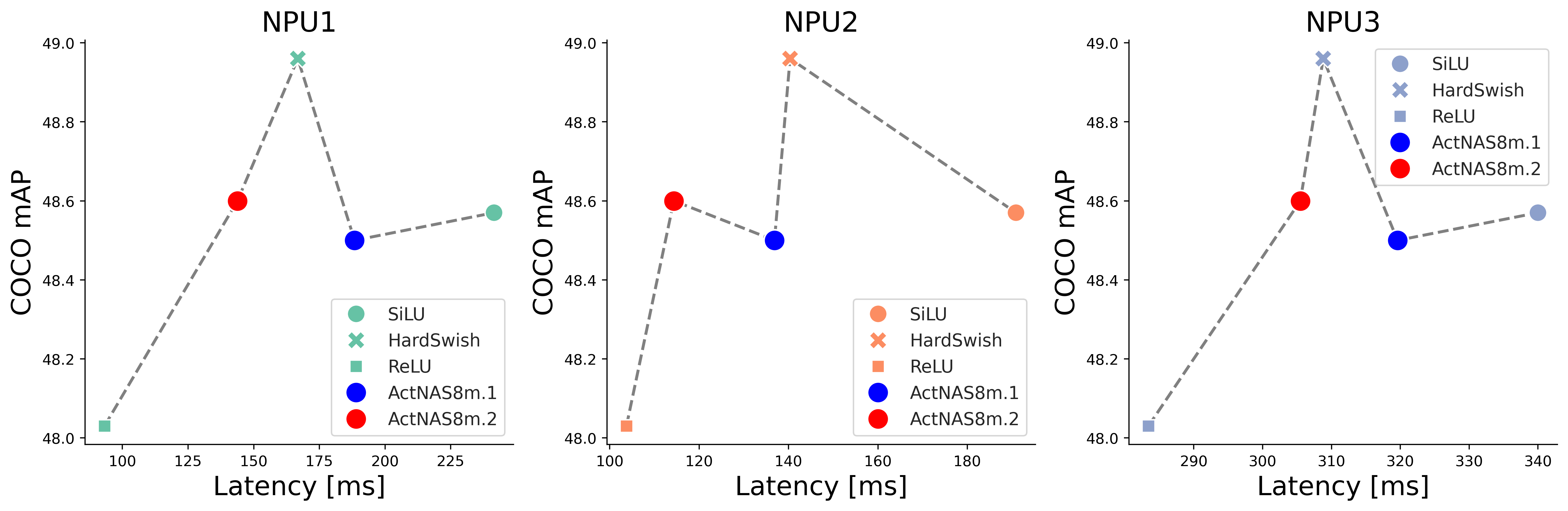}
    \caption{Performance of YOLO8m reference models and ActNAS models on different edge NPUs.}
    \label{fig:actnas_results_yolov8}
\end{figure*}

{ActNAS5n.1} and {ActNAS5n.2} were generated using our activation NAS approach, targeting the NPU as the primary device, as discussed in the previous section. The mAP of {ActNAS5n.1} model is within 1\% drop compared to both SiLU and HardSwish variants but this model has lowest latency showing 6.26\%, 3.32\%, 0.97\% and 1.69\% improvements in efficiency on the NPU1, Jetson Nano GPU, Cortex A57 CPU and A53 CPU, respectively, compared to the Hardswish model. It also outperformed the SiLU model by 22.28\%, 8.90\%, 11.92\%, and 13.64\% on the these devices. Furthermore, we found that {ActNAS5n.1} required 58\% less memory than the SiLU model on NPU1. {ActNAS5n.2}, generated with the Jetson Nano GPU as the target, outperformed both the baseline and HardSwish models on Jetson Nano, highlighting the hardware-aware nature of our search method, as it adapts to different hardware profiles.
We also observed that the best latency-accuracy trade-off models tends to use ReLU variants (ReLU6, LeakyReLU, and ReLU) for initial layers. None of the ActNAS models utilized SiLU or Hardswish in the first two layers. This suggests that replacing the initial layers with more efficient activations and following them with high-performance activations selected via NAS to minimize accuracy loss produces optimal models across a range of hardware devices. Figure \ref{fig:results_plot} compares our models with baseline models, showing that Activation NAS models offer the best accuracy-latency trade-off on all tested hardware.
In order to test the scalability of this approach, we repeated these experiments with YOLO8m models which has more activation layers compared to YOLO5n. We observerd that the impact of different activations is more prominent on NPUs as latency is not only affected by the complexity of operations but also by the memory usage. Therefore, we focused all the YOLO8m experiments on three different NPUs. Table \ref{table:table_npus_yolov8} shows the mAP/latency trade-off of {ActNAS8m.1} and {ActNAS8m.2} models compared to baseline models. These models are created to be NPU2 hardware aware and it is evident that {ActNAS8m.2} model is 1.66x faster compared to the SiLU model without significant impact on accuracy of the model. Figure \ref{fig:actnas_results_yolov8} shows the mAP-latency curve for ActNAS YOLO8m models compared to baseline model.

\section{Hardware Awareness}
\label{hw_awareness}

\begin{figure}
    \centering
    \includegraphics[width=1.0\linewidth]{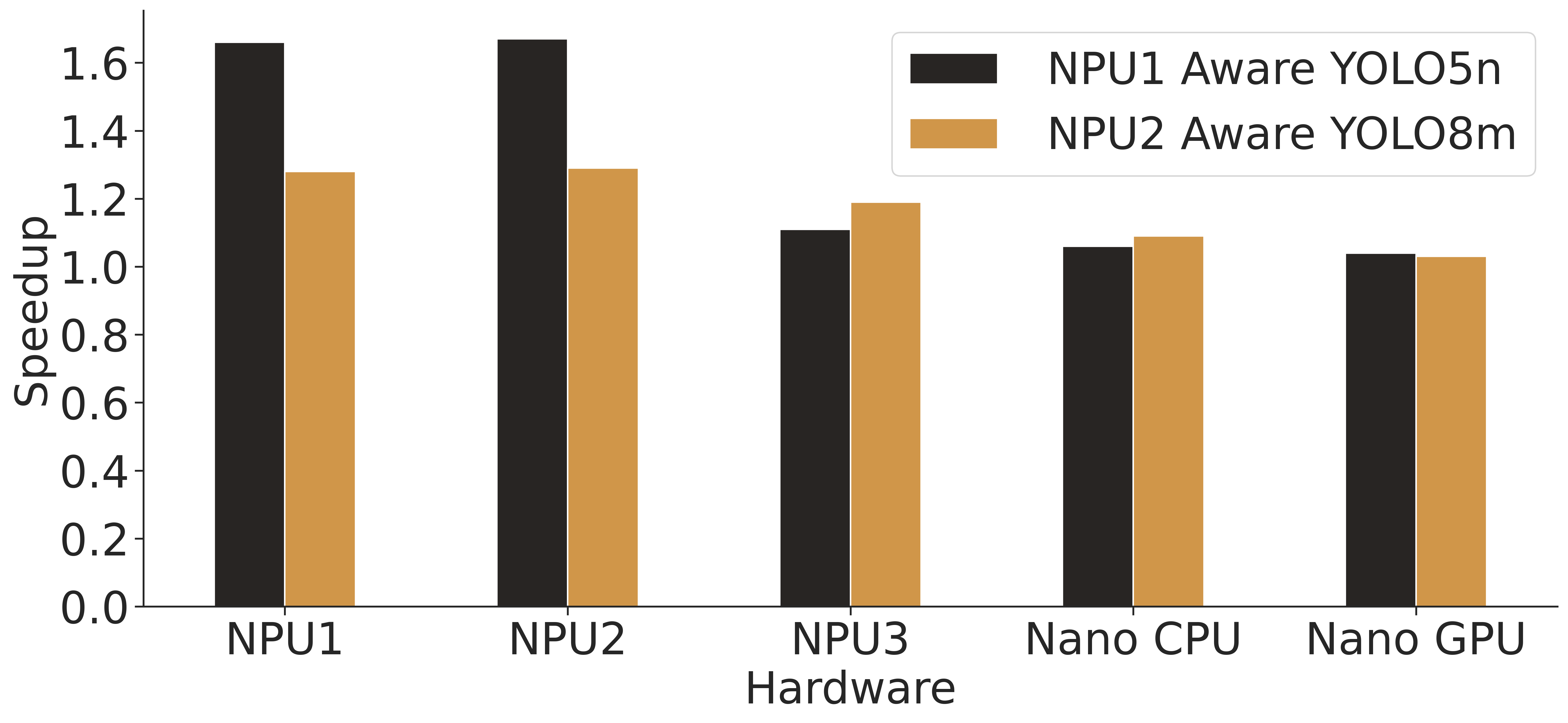}
    \caption{ActNAS model speedup on different hardware where the model is optimized for NPU1 (NPU1 and NPU2 have similar configuration)}
    \label{fig:hw_aware}
\end{figure}

As mentioned in section \ref{methodology}, all of the ActNAS models use one of the edge device for latency/memory computation. Since the latency and memory impact of each change in the model correlates to the hardware architecture, memory limitations and compiler optimization, we expect the resulting optmized model to have the best performance on that hardware. In other words, the model is "device" aware. For example, {ActNAS5n.1}, {ActNAS5n.2}, {ActNAS8m.1} and {ActNAS8m.2} are NPU1, Jetson Nano GPU, NPU2 and NPU2 aware models. The bar chart in Figure \ref{fig:hw_aware} shows the speedup of NPU1 aware {ActNAS5n.1} and NPU2 aware {ActNAS8m.2} speed up on five different devices. NPU1 and NPU2 have similar architecture but with different memory configuration. We can see that both NPU aware models provide highest speedup on these devices but very low speedup on other devices. These results also suggest that the NAS is able to learn model configuration suited best for these devices. 

\section{Conclusion}
\label{conclusion}
 
In this paper, we present a novel approach called Activation NAS (ActNAS), designed to generate YOLO model architectures through a unique search space of varied activation functions for each layer. Our mixed-activation models deliver performance comparable to SiLU/Hardswish-based models on the COCO dataset, while offering up to 1.66x greater efficiency, depending on the hardware. This efficiency is achieved using a Neural Architecture Search (NAS) approach that leverages the zero-cost NWOT score, enabling us to skip the costly process of training a vast number of candidate models, thus making model optimization more practical. Our experiments suggest that, rather than using a single activation function across all model layers, a combination of latency and accuracy constraints can guide the creation of custom mixed-activation models tailored to meet specific accuracy, latency, and memory requirements. Furthermore, we demonstrate that models optimized with particular hardware in mind (hardware-aware models) achieve peak performance on that hardware, yet may underperform on others. This finding is crucial for model accelerators and NPUs with varying architectures, where hardware-agnostic optimizations often fail to yield efficient models.

{
    \small
    \bibliographystyle{ieeenat_fullname}
    \bibliography{main}

\begin{thebibliography}{28}
\providecommand{\natexlab}[1]{#1}
\providecommand{\url}[1]{\texttt{#1}}
\expandafter\ifx\csname urlstyle\endcsname\relax
  \providecommand{\doi}[1]{doi: #1}\else
  \providecommand{\doi}{doi: \begingroup \urlstyle{rm}\Url}\fi

\bibitem[Abdelfattah et~al.(2021)Abdelfattah, Mehrotra, Dudziak, and Lane]{abdelfattah2021zero}
Mohamed~S Abdelfattah, Abhinav Mehrotra, {\L}ukasz Dudziak, and Nicholas~D Lane.
\newblock Zero-cost proxies for lightweight nas.
\newblock \emph{arXiv preprint arXiv:2101.08134}, 2021.

\bibitem[Agarap(2018)]{relu}
AF Agarap.
\newblock Deep learning using rectified linear units (relu).
\newblock \emph{arXiv preprint arXiv:1803.08375}, 2018.

\bibitem[Bodyanskiy and Kostiuk(2022)]{Adaptive_hybrid_activation}
Yevgeniy Bodyanskiy and Serhii Kostiuk.
\newblock Adaptive hybrid activation function for deep neural networks.
\newblock \emph{System research and information technologies}, pages 87--96, 2022.

\bibitem[Elfwing et~al.(2018)Elfwing, Uchibe, and Doya]{silu}
Stefan Elfwing, Eiji Uchibe, and Kenji Doya.
\newblock Sigmoid-weighted linear units for neural network function approximation in reinforcement learning.
\newblock \emph{Neural networks}, 107:\penalty0 3--11, 2018.

\bibitem[Everingham et~al.(2012)Everingham, Van~Gool, Williams, Winn, and Zisserman]{pascal-voc-2012}
M. Everingham, L. Van~Gool, C.~K.~I. Williams, J. Winn, and A. Zisserman.
\newblock The {PASCAL} {V}isual {O}bject {C}lasses {C}hallenge 2012 {(VOC2012)} {R}esults, 2012.

\bibitem[Glenn(2022)]{yolov6_ultralyticsv5}
Jocher Glenn, 2022.

\bibitem[Howard et~al.(2019)Howard, Sandler, Chu, Chen, Chen, Tan, Wang, Zhu, Pang, Vasudevan, et~al.]{hardswish}
Andrew Howard, Mark Sandler, Grace Chu, Liang-Chieh Chen, Bo Chen, Mingxing Tan, Weijun Wang, Yukun Zhu, Ruoming Pang, Vijay Vasudevan, et~al.
\newblock Searching for mobilenetv3.
\newblock In \emph{Proceedings of the IEEE/CVF international conference on computer vision}, pages 1314--1324, 2019.

\bibitem[Jocher(2020)]{Ultralytics}
Glenn Jocher.
\newblock Ultralytics yolov5, 2020.

\bibitem[Lazarevich et~al.(2023)Lazarevich, Grimaldi, Kumar, Mitra, Khan, and Sah]{yolobench}
Ivan Lazarevich, Matteo Grimaldi, Ravish Kumar, Saptarshi Mitra, Shahrukh Khan, and Sudhakar Sah.
\newblock Yolobench: Benchmarking efficient object detectors on embedded systems, 2023.

\bibitem[Li et~al.(2023{\natexlab{a}})Li, Li, Geng, Jiang, Cheng, Zhang, Ke, Xu, and Chu]{yolov6}
Chuyi Li, Lulu Li, Yifei Geng, Hongliang Jiang, Meng Cheng, Bo Zhang, Zaidan Ke, Xiaoming Xu, and Xiangxiang Chu.
\newblock Yolov6 v3. 0: A full-scale reloading.
\newblock \emph{arXiv preprint arXiv:2301.05586}, 2023{\natexlab{a}}.

\bibitem[Li et~al.(2023{\natexlab{b}})Li, Yang, Bhardwaj, and Marculescu]{li2023zico}
Guihong Li, Yuedong Yang, Kartikeya Bhardwaj, and Radu Marculescu.
\newblock Zico: Zero-shot nas via inverse coefficient of variation on gradients.
\newblock \emph{arXiv preprint arXiv:2301.11300}, 2023{\natexlab{b}}.

\bibitem[Lin et~al.(2014)Lin, Maire, Belongie, Bourdev, Girshick, Hays, Perona, Ramanan, Zitnick, and Dollár]{lin2014microsoft}
Tsung-Yi Lin, Michael Maire, Serge Belongie, Lubomir Bourdev, Ross Girshick, James Hays, Pietro Perona, Deva Ramanan, C.~Lawrence Zitnick, and Piotr Dollár.
\newblock Microsoft coco: Common objects in context, 2014.
\newblock cite arxiv:1405.0312.

\bibitem[Lite(2024)]{tflite}
TensorFlow Lite.
\newblock Tensorflow lite for mobile and edge, 2024.
\newblock Accessed: 2024-07-16.

\bibitem[Martins and Astudillo(2016)]{sparsemax}
Andre Martins and Ramon Astudillo.
\newblock From softmax to sparsemax: A sparse model of attention and multi-label classification.
\newblock In \emph{International conference on machine learning}, pages 1614--1623. PMLR, 2016.

\bibitem[Mellor et~al.(2021)Mellor, Turner, Storkey, and Crowley]{mellor2021neural}
Joe Mellor, Jack Turner, Amos Storkey, and Elliot~J Crowley.
\newblock Neural architecture search without training.
\newblock In \emph{International Conference on Machine Learning}, pages 7588--7598. PMLR, 2021.

\bibitem[Mitchell et~al.(2022)Mitchell, Consulting, O’Sullivan, and Dunning]{PuLP}
Stuart Mitchell, Stuart~Mitchell Consulting, Michael O’Sullivan, and Iain Dunning.
\newblock Pulp: A linear programming toolkit for python, 2022.

\bibitem[Nguyen and Mondelli(2020)]{NAS-Pyramidal-Topology}
Quynh~N Nguyen and Marco Mondelli.
\newblock Global convergence of deep networks with one wide layer followed by pyramidal topology.
\newblock \emph{Advances in Neural Information Processing Systems}, 33:\penalty0 11961--11972, 2020.

\bibitem[ONNX(2024)]{onnx}
ONNX.
\newblock Open neural network exchange, 2024.
\newblock Accessed: 2024-07-16.

\bibitem[Pearce et~al.(2021)Pearce, Brintrup, and Zhu]{softmax}
Tim Pearce, Alexandra Brintrup, and Jun Zhu.
\newblock Understanding softmax confidence and uncertainty.
\newblock \emph{arXiv preprint arXiv:2106.04972}, 2021.

\bibitem[Privietha and Raj(2022)]{Hybrid_Activation}
P Privietha and V~Joesph Raj.
\newblock Hybrid activation function in deep learning for gait analysis.
\newblock In \emph{2022 International Virtual Conference on Power Engineering Computing and Control: Developments in Electric Vehicles and Energy Sector for Sustainable Future (PECCON)}, pages 1--7. IEEE, 2022.

\bibitem[Ramachandran et~al.(2017)Ramachandran, Zoph, and Le]{search_activation}
Prajit Ramachandran, Barret Zoph, and Quoc~V Le.
\newblock Searching for activation functions.
\newblock \emph{arXiv preprint arXiv:1710.05941}, 2017.

\bibitem[Sah et~al.(2024)Sah, Ganji, Grimaldi, Kumar, Hoffman, Rohmetra, and Saboori]{sah2024mcubench}
Sudhakar Sah, Darshan~C Ganji, Matteo Grimaldi, Ravish Kumar, Alexander Hoffman, Honnesh Rohmetra, and Ehsan Saboori.
\newblock Mcubench: A benchmark of tiny object detectors on mcus.
\newblock \emph{arXiv preprint arXiv:2409.18866}, 2024.

\bibitem[Terven et~al.(2023)Terven, C{\'o}rdova-Esparza, and Romero-Gonz{\'a}lez]{yolo_survey}
Juan Terven, Diana-Margarita C{\'o}rdova-Esparza, and Julio-Alejandro Romero-Gonz{\'a}lez.
\newblock A comprehensive review of yolo architectures in computer vision: From yolov1 to yolov8 and yolo-nas.
\newblock \emph{Machine Learning and Knowledge Extraction}, 5\penalty0 (4):\penalty0 1680--1716, 2023.

\bibitem[Trauth~Jr and Woolsey(1969)]{trauth1969integerlinear}
CA Trauth~Jr and RE Woolsey.
\newblock Integer linear programming: a study in computational efficiency.
\newblock \emph{Management Science}, 15\penalty0 (9):\penalty0 481--493, 1969.

\bibitem[Wang et~al.(2024)Wang, Chen, and Liu]{YOLOv10}
Ao Wang, Hui Chen, and Lihao Liu.
\newblock Yolov10: Real-time end-to-end object detection.
\newblock \emph{arXiv preprint arXiv:2405.14458}, 2024.

\bibitem[Xu(2015)]{leakyrelu}
Bing Xu.
\newblock Empirical evaluation of rectified activations in convolutional network.
\newblock \emph{arXiv preprint arXiv:1505.00853}, 2015.

\bibitem[Zhu et~al.(2022)Zhu, Liu, Chrysos, and Cevher]{NAS-skip-connection}
Zhenyu Zhu, Fanghui Liu, Grigorios Chrysos, and Volkan Cevher.
\newblock Generalization properties of nas under activation and skip connection search.
\newblock \emph{Advances in Neural Information Processing Systems}, 35:\penalty0 23551--23565, 2022.

\bibitem[Zoph(2016)]{NAS-original}
B Zoph.
\newblock Neural architecture search with reinforcement learning.
\newblock \emph{arXiv preprint arXiv:1611.01578}, 2016.

\end{thebibliography}
}


\end{document}